\documentclass[letterpaper]{article} 
\usepackage[submission]{aaai}  
\usepackage{times}  
\usepackage{helvet}  
\usepackage{courier}  
\usepackage[hyphens]{url}  
\usepackage{graphicx} 
\usepackage{natbib}  
\usepackage{caption} 
\usepackage{algorithm}
\usepackage{algorithmic}

\usepackage{newfloat}
\usepackage{listings}
\DeclareCaptionStyle{ruled}{labelfont=normalfont,labelsep=colon,strut=off} 
\floatstyle{ruled}
\newfloat{listing}{tb}{lst}{}
\floatname{listing}{Listing}
\usepackage{graphicx}
\usepackage{subfigure}
\usepackage{amssymb}
\usepackage{amsmath, amsthm}
\usepackage{booktabs}
\usepackage{multirow}
\usepackage{makecell}
\usepackage{IEEEtrantools}
\usepackage{fge}
\usepackage{ebproof}
\usepackage{footmisc}
\usepackage[inline]{enumitem}
\usepackage{microtype}
\usepackage{latexsym}
\usepackage{mathtools}
\usepackage{xcolor, color}

\newtheorem{proposition}{Proposition}

\title{Unlearning with Fisher Masking}
\author {
    Yufang Liu,
    Changzhi Sun, 
    Yuanbin Wu, 
    Aimin Zhou  \\
}
\affiliations {
    School of Computer Science and Technology \\
    East China Normal University \\
    yfliu.antnlp@gmail.com, czsun.cs@gmail.com, \{ybwu, amzhou\}@cs.ecnu.edu.cn
}

\begin{document}
\maketitle              
%

\begin{abstract}
  Machine unlearning aims to revoke some 
  training data after learning
  in response to requests from users,
  model developers, and administrators.
  Most previous methods are based on direct fine-tuning,
  which may neither remove data completely nor 
  retain full performances on the remain data.
  In this work, we find that, by first masking some 
  important parameters before fine-tuning,
  the performances of unlearning could be significantly improved.
  We propose a new masking strategy tailored to unlearning
  based on Fisher information.
  Experiments on various datasets and network structures 
  show the effectiveness of the method:
  without any fine-tuning, the proposed Fisher masking 
  could unlearn almost completely while maintaining 
  most of the performance on the remain data. 
  It also exhibits stronger stability compared 
  to other unlearning baselines.
  
\end{abstract}

\section{Introduction}

Machine learning algorithms need data for building models.
As a large amount of data-driven models are rushing into
people's daily life, operations on regularizing data usage
become crucial. One such operation is 
removing data from deployed models (also called \emph{machine unlearning} \cite{DBLP:conf/sp/CaoY15}).
For instance, legal laws (e.g., 
General Data Protection Regulation (GDPR), 
California Consumer Privacy Act (CCPA)
and Personal Information Protection and Electronic Documents Act (PIPEDA))
declare that users have the right to ask business companies to
revoke their personal data.
At the same time, models can benefit from
removing wrongly annotated data
\cite{DBLP:conf/smc/RajmadhanGH17,DBLP:conf/cvpr/0002HWZ21,DBLP:conf/cvpr/PangZQJ21},
systematic biases \cite{DBLP:conf/emnlp/ZhaoC20,DBLP:conf/cvpr/KimKKKK19,DBLP:conf/icpr/SernaPMF20},
and backdoor poisoned data \cite{DBLP:journals/ijon/ChenD21,DBLP:conf/aaai/YanLTWLCP21,DBLP:conf/acl/QiLCZ0WS20}.

\begin{figure}[tbp]
    \centering 
    \includegraphics[scale=0.2]{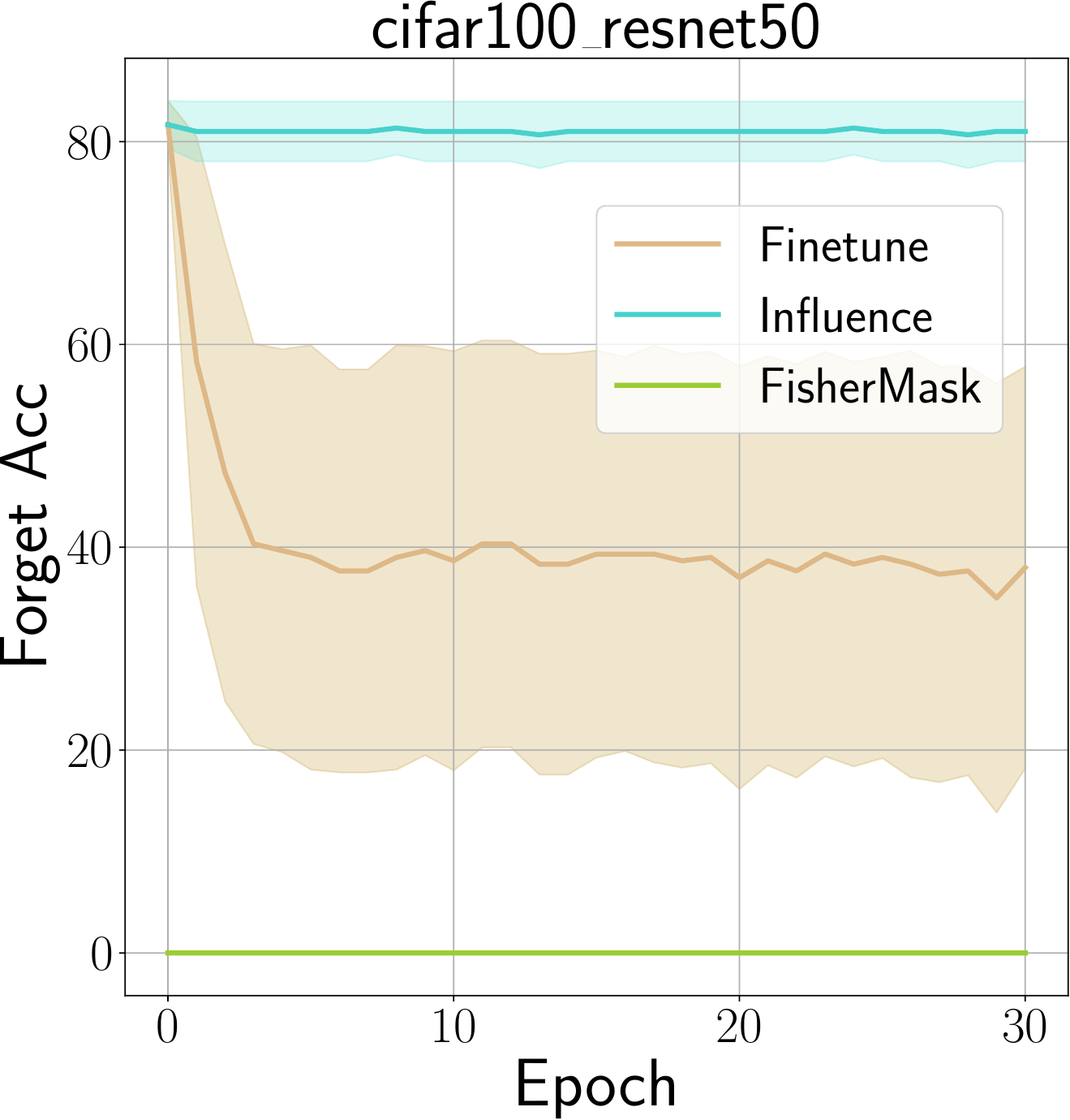}
    \caption{Example unlearning performances. 
    We train ResNet50 with CIFAR-100, and then ask algorithms to 
    remove all pictures from a class. 
    y-axis is the testing performance on that removed class
    during unlearning.
    The results indicate that both 
    direct fine-tuning and influence function 
    can not effectively unlearn data points.
    FisherMask is our proposed method.}
    \label{fig:influence}
\end{figure}

Given the training set and a subset to remove, 
the straightforward (and the optimal) way of 
unlearning is re-learning the model.
It guarantees a clean removal, but the computation
cost is high.
More computationally efficient approaches
are based on fine-tuning:
starting a new learning process on the current model
with only remain data. 
It is known that as the fine-tuning process proceeds,
the model gradually forgets those unseen data points 
(\emph{catastrophic forgetting} \cite{kirkpatrick2017overcoming}).
However, fine-tuning-based unlearning 
could be slow and incomplete in practice.
For example, in Figure \ref{fig:influence},
we ask ResNet50 to remove all pictures belong to one class of CIFAR-100,
and after fine-tuning on remain samples, it still has about $40\%$
accuracy on the removed class (a clean removal should be $0$).
Another widely studied fine-tuning strategy 
is based on second-order optimization.
\cite{DBLP:conf/icml/KohL17,DBLP:conf/icml/GuoGHM20} 
show that with a one-step Newton update
(also called \emph{influence function}),
the new model's prediction behaviour 
correlates well with the 
re-learned model, but the strong correlation there
does not imply a successful unlearning:
in Figure \ref{fig:influence}, 
the one-step Newton update almost remove no information
($80\%$ accuracy).
Hence, given the special initial state (parameters), 
unlearning with fine-tuning is still a challenge:
it is hard to escape the local optimum of the old model.

In this work, we study methods to accelerate unlearning
by adding proper perturbations on its initial state.
Instead of adding them randomly 
(as suggested by most Hitchhiker's guides to escape local optimum),
we would like the perturbations are biased towards
the task of removing data.
To accomplish this, we first identify parameters which are important 
for modeling the excluded data, and 
mask them before fine-tuning the model.
Our main finding is that Fisher information plays
a key role in masking parameters:
it characterizes how parameters contribute to 
the distance between models before and after unlearning.
We develop a new masking strategy based on Fisher information
which shows strong unlearning performances across different datasets
and deep neural network structures.
We conduct extensive empirical evaluations on fine-tuning-based
unlearning methods with fair and reproducible configurations
The main empirical findings are,
\begin{itemize}[leftmargin=*]
    \item compared with direct fine-tuning,
    masking strategies can significantly improve unlearning performance.
    In fact, even with random masking, the unlearning process could be 
    accelerated.
    \item comparing with neuron activation scores,
    masking with Fisher information is effective on 
    balancing removing and reserving.
    Without any fine-tuning, Fisher masking 
    could unlearn almost completely while maintaining most of the 
    performance on the remaining data. 
    \item unlearning algorithms could be unstable with respect to 
    different experiment settings and random seeds.
    The Fisher masking exhibits the best
    stability among current fine-tuning-based methods.
\end{itemize}

\section{Related Work}

Machine unlearning, first proposed by 
\citet{DBLP:conf/sp/CaoY15} 
in the context of statistical query learning, aims to forget training samples
for data protection and model security. 
Most of the previous works in machine unlearning focus on linear models, 
need to calculate the reverse of hessian matrix, 
and perform a single SGD update towards the minimizer of the approximation \cite{DBLP:conf/icml/KohL17,DBLP:conf/icml/GuoGHM20,DBLP:conf/aistats/IzzoSCZ21}.
Besides linear models, 
\citet{DBLP:conf/nips/GinartGVZ19}
investigate an effective data deletion algorithm for the specific setting 
of k-means clustering. 
\citet{DBLP:conf/icml/BrophyL21}
apply a variant of random forests that 
enables the removal of training data with minimal retraining.
For deep neural networks, 
\citet{DBLP:conf/cvpr/GolatkarAS20} 
try to add a fisher noise 
to hide the information about unlearn data.
The work closest to ours is \cite{DBLP:conf/www/Wang0XQ22} (which is a concurrent work).
They try to scrub memories for each category in federated learning.
Unlike our method, they calculate activation maps on the dataset in each layer 
and use TF-IDF to choose neurons after grouping and averaging activation maps 
by category. After pruning, there also utilize a fine-tuning process to recover the performance.

Contrary to machine unlearning, life-long learning or continual learning, 
is often viewed as the concept to learn many tasks sequentially without forgetting
the knowledge obtained from preceding tasks. The term ``forgetting" mentioned here
is \emph{Catastrophic Forgetting} \cite{DBLP:conf/nips/French93}, 
which results in model overfitting on the currently available data 
and suffering from performance deterioration on the previously trained data.
However, 
\citet{DBLP:conf/cvpr/GolatkarAS20} 
show that finetune on the remain dataset
from the original trained model could not suffer catastrophic forgetting,
while our experiments present different results which we attribute 
to the different learning settings.
A lot of works have done to constraint forgetting, 
such as, regularization-based methods \cite{kirkpatrick2017overcoming,DBLP:conf/eccv/AljundiBERT18} propose to selectively
slow down the learning rate of task important parameters;
rehearsal-based methods \cite{DBLP:conf/iclr/ChaudhryRRE19,DBLP:conf/icra/HayesCK19} 
save a data buffer to recover performance of old data when new task comes;
and architecture-based methods \cite{DBLP:conf/icml/LiZWSX19,DBLP:conf/iclr/LooST21}
have separate components for each task.
Our work based on the previous found \cite{DBLP:journals/pnas/BauZSLZ020} 
that one subset of neurons can be highly activated by specific training images,
which motivates us to separate the parameters for unlearn data and remain data.

Differential privacy \cite{DBLP:journals/fttcs/DworkR14}
is another related task, and it focuses on guaranteeing that 
information about training data could not be leaked by the trained model. 
It requires the model to remain unchanged after replacing 
any one of the training data. 
Machine unlearning could be viewed as a generalization of differential privacy
when removing one data point according to definition.
Previous studies \cite{DBLP:conf/ccs/AbadiCGMMT016,DBLP:journals/corr/abs-2111-00173}
try to add noise on gradient during learning to hide the data information,
which will inevitably degrade model performance and could not be applied 
for scenarios with high accuracy requirements, 
such as face recognition and financial risk measurement.

\section{Background}

Given a dataset $D = \{(x_i, y_i)\}_{i=1}^{|D|}$, 
$y_i \in \mathcal{Y}$ is 
the label of input $x_i \in \mathcal{X}$.
A learning algorithm tries to minimize the loss function 
$\mathcal{L}(w, D) = \sum_{i=1}^N \ell(x_i, y_i, w)$ on the training set $D$,
where $w$ is the model parameter, and $\ell$ is the log-loss 
$\ell(x, y, w) = -\log p(y|x, w)$.
We denote $w^* = \arg\min_w \mathcal{L}(w, D)$.

Let $D_f$ be the subset that we want to remove from 
the model (\emph{forget set}),
and $D_r = D\setminus D_f$ contains remain data samples
(\emph{remain set}).
$D_f$ could be any subset of $D$, 
but the retrained model after removing a random group 
could be identical to the original model due to the fact 
that there are usually many copies of each data in the dataset.
For evaluation purpose,
previous works focus on cases that $D_f$ contains 
all samples belong to the same class (remove a whole category)
\cite{DBLP:conf/ijcai/0001IIM21,DBLP:conf/www/Wang0XQ22}.
In this case, the target of unlearning 
is to obtain $\hat{w}_r$ which 1) has similar performances 
on $D_r$ as $w_r^* = \arg\min_w \mathcal{L}(w, D_r)$,
and 2) contains no information about $D_f$ 
(i.e., zero accuracy on samples in $D_f$ like $w_r^*$).
\footnote{For arbitrary $D_f$, zero accuracy is not
sufficient for validating unlearning.
For example, if we have $D_f = D_r$ (duplicate datasets), 
the unlearned model should have
identical behaviours on $D_f$ and $D_r$.
We will evaluate the performance of arbitrary $D_f$
in the task of denoise (Table \ref{table:outlier_deletion}).}
Besides removing the whole category, we also conduct experiments on removing poisoned samples and wrongly labeled samples to test the unlearning performance on random group.


Instead of fully re-training from scratch,
one could solve the objective function
$\arg\min_w \mathcal{L}(w, D_r)$ from 
$w^*$ (i.e., fine-tuning).
While any optimization procedure could be applied
(e.g., SGD), 
 \citet{DBLP:conf/icml/KohL17,DBLP:conf/nips/KohATL19,DBLP:conf/cvpr/GolatkarAS20}
show that for the special setting of the initial state 
($w_0 = w^*$),
one-step Newton update could be an effective move
towards the unlearning target $w_r^*$.
Specifically, the influence function used in \cite{DBLP:conf/icml/KohL17,DBLP:conf/nips/KohATL19}
approximates the difference between $w^*$ and $w_r^*$ with,
\begin{IEEEeqnarray*}{c}
    w_r^* \approx \hat{w}_r = w^* + \frac{1}{|D_f|} 
    \nabla_{w}^2\mathcal{L}(w^*, D)^{-1}
    \nabla_{w}\mathcal{L}(w^*, D_f).
\end{IEEEeqnarray*}
The approximation is obtained by Taylor expansion
of $\mathcal{L}(w, D)$ at the stationary $w^*$,
which assumes a small forget set $D_f$.
\citet{DBLP:conf/cvpr/GolatkarAS20} 
add a noise term to the one-step Newton update
which aims to approximately minimize 
KL-divergence between $\hat{w}_r$ and $w_r^*$,
\begin{IEEEeqnarray*}{rl}
    w_r^* \approx \hat{w}_r = w^* - 
    \nabla_{w}^2\mathcal{L}(w^*, D_r)^{-1}
    \nabla_{w}\mathcal{L}(w^*, D_r) \\
    + (\lambda \sigma^2)^{\frac{1}{4}} 
    \nabla_{w}^2\mathcal{L}(w^*, D_r)^{-\frac{1}{4}}\epsilon,
    \label{eq:fishernoise1}
\end{IEEEeqnarray*}
where $\lambda, \sigma$ are hyperparameters,
and $\epsilon \thicksim N(0, I)$ is a Gaussian noise.

Although fine-tuning with Newton-updates only needs one step,
it is expensive to compute Hessian matrices for deep neural networks.
\citet{DBLP:conf/icml/KohL17} 
apply the iterative LiSSA algorithm \cite{JMLR:v18:16-491}
to approximate the Hessian.
\citet{DBLP:conf/cvpr/GolatkarAS20} 
simply drop the second term of Equation \ref{eq:fishernoise1},
and approximate Hessian with diagonals
of Fisher matrix in the noise term,
\begin{IEEEeqnarray}{c}
    \hat{w}_r = w^* + (\lambda \sigma^2)^{\frac{1}{4}} h^{-\frac{1}{4}},
    \label{eq:fishernoise2}
\end{IEEEeqnarray}
where vector $h$ contains diagonal entries of the Fisher matrix 
computed on $w^*$ for dataset $D_r$.

\section{Unlearning Approaches}


In this section, we first show that Fisher information 
is important for identifying key parameters for unlearning,
and based on this observation, we propose a new masking strategy
(\texttt{FisherMask}). We then describe an alternative masking method
(\texttt{ActivationMask}) 
in Section \ref{ActivationMask}.
We also introduce a setting of learning rates 
for the following fine-tuning process which makes
unlearning stable in practice.

\subsection{Masking with Fisher Information}
\label{FisherMask}

For a distribution $p(y|x, w)$,
Fisher matrix (and its empirical estimation) 
is defined by
\begin{IEEEeqnarray*}{c}
F\triangleq \mathrm{E}_{x,y}
\left[\nabla_w\log p(y|x, w)\nabla_w\log p(y|x, w)^T\right] \\
\approx
\frac{1}{|D|}\sum_{i=1}^{|D|}
\nabla_w\log p(y_i|x_i, w)\nabla_w\log p(y_i|x_i, w)^T.
\end{IEEEeqnarray*}
For large-scale neural networks,
it is usually expensive to use full $F$,
thus we will further approximate $F$ with its diagonal
$\mathrm{diag}(F)$
following \cite{kirkpatrick2017overcoming,DBLP:conf/cvpr/GolatkarAS20}.
It is known that $F$ equals to 
negative expectation of $\log p(y|x, w)$'s Hessian,
\begin{IEEEeqnarray*}{c}
F =-\mathrm{E}_{x, y}\nabla_w^2\log p(y|x, w).
\end{IEEEeqnarray*}

We now take linear regression as an example
to show the role of Fisher information in unlearning.
Let $p(y|x, w) = \frac{1}{Z} \exp\{-\frac{1}{2}(w^Tx - y)^2\}$,
and $X = [x_1, x_2, \ldots, x_{|D|}]$.
The empirical Fisher now is $F = |D|^{-1} XX^T$.
Let 
$F_{jj} = \sum_{x_i\in D} x_{ij}^2$,
be the diagonals of the Fisher matrix 
(to simplify notations the factor $|D|$ is dropped),
where $x_{ij}$ is the $j$-th dimension of $x_i$.
Let $F_{r, jj} = \sum_{x_i\in D_r} x_{ij}^2$
and 
$F_{f, jj} = \sum_{x_i\in D_f} x_{ij}^2$
be the remain set and forget set's contribution to 
the $j$-th diagonal of the Fisher matrix.

Let $M$ be the set of parameters 
to be masked,
and $\hat{w}_r$ be the parameter obtained by masking
$w^*$ with $M$, whose $j$-th entry is
$\hat{w}_{r,j} = \big\{ \begin{smallmatrix} 
w^*_j, & j\notin M \\ 0, & j\in M \end{smallmatrix}$.
\begin{proposition}
\label{thm:upperbound}
For linear regression, 
if we approximate Fisher matrix $F$ with its diagonals
$\mathrm{diag}(F)$
and assume all diagonals are restrict positive,
the KL-divergence between the optimal model $w_r^*$ and
the masked model $\hat{w}_r$ has the following upper bound,
\begin{IEEEeqnarray}{c}
\label{eq:upperbound}
\mathrm{KL}(w_r^*, \hat{w}_r) \! \leq \! \frac{\lambda}{2|D|} 
\left(
c \! + \! 2c_1\sum_{j\notin M} \frac{1}{F_{jj}^2}
\left(\frac{F_{f, jj}}{F_{r, jj}}\right)^2
\right),
\end{IEEEeqnarray}
where $\lambda$ is the largest eigenvalue of $XX^T$,
and $c, c_1$ are constants depend on the remain set $D_r$. 
See appendix \ref{sec:proof} for a derivation.
\end{proposition}

The upper bound implies that to make the masked 
parameter $\hat{w}_r$ close to the unlearning target $w^*_r$, 
the unmasked set $\overline{M}$ should 
contain those parameters with 
small Fisher information contribution on the forget set $F_{f, jj}$ and 
large Fisher information contribution on the remain set $F_{r, jj}$,
which means the masking strategy should do the opposite.
Therefore, we develop \texttt{FisherMask} strategy to select top $R$
parameters according to $F_{f, jj}-F_{r, jj}$ as $M$.


Proposition \ref{thm:upperbound} could be extended
to generalized linear models:
the proof depends on the close form solution of linear 
regression, while similar estimation of solutions
could be established for 
generalized linear models \cite{NIPS2015_17d63b16}.
We also remark that, though \texttt{FisherMask} performs quite
well for deep models (e.g., models with parameterized 
representation layers), 
we now don't obtain a similar upper bound for them.


\subsection{Masking with Activation Values}
\label{ActivationMask}

In neural networks,
an alternative way to measure importance of parameters
is inspecting activation states of their corresponding 
neurons.
As 
\citet{erhan2009visualizing} suggested,
maximizing a neuron's activation value with respect to input
could be a good first-order representation 
of what the neuron is doing. 
Here, we could find neurons which maximize the activation 
on the forget set $D_f$
and mask corresponding parameters
to get perturbations on the old model.

Suppose we have a trained $L$ layer CNN model.
First, for each training sample $i$, 
we average activation scores of an output channel $j$
(obtained by Conv-BatchNorm-ReLu operations 
on an intermediate input channel),
and record the score to $A_{ij}$ of a table $A$.
Then, for each channel, we can compute its
averaged activation values
$A_{D_r, j} = \frac{1}{|D_r|}\sum_{i\in D_r}A_{ij}$ over the remain
set $D_r$,
and similarly, $A_{D_f, j}$ on the forget set $D_f$.
The \texttt{ActivationMask} strategy identifies 
top $R$ of channels with large
$A_{D_f, j} - A_{D_r, j}$,
and mask CNN kernel parameters connecting with
those channels.




An improved version of \texttt{ActivationMask}
is proposed by \cite{DBLP:conf/www/Wang0XQ22} which not only 
looks at how a channel activates,
but also how it contributes to the whole activation pattern of 
the entire class 
(their method can only remove all samples of a class).
Specifically, their \texttt{TF-IDF} method
masks neurons with
term-frequency inverse-document-frequency scores,
which analogize channels to words and classes to documents
in information retrieval.
It is worth a mention that, 
the activation value there is calculated before 
going through BatchNorm layer,
which means information stored in BatchNorm layers
can not be removed.

\subsection{Fine-Tuning}
\label{finetune}
After masking, we fine-tune parameters on the $D_r$ to recover the performance on remain data.
We find that the final unlearning performances
could be sensitive to different settings of learning rate,
which are usually ignored in current unlearning configurations. \footnote{
Appendix \ref{sec:learning_rate} includes the experiment on different learning rate schedule.
}
For example, 
\citet{DBLP:conf/www/Wang0XQ22} choose a
fixed learning rate $0.1$ in fine-tuning process 
(and in training),
but constant learning rate is not the standard 
setting of modern optimization algorithms.

In experiments, we deploy a learning rate scheduler for 
unlearning to mimic
the original learning process in 
a shorter period (denoted by $S$, and we set $S=5$).
For example,
if the original learning process
triggers a decay of rate at $1/2$
of learning epochs, then the unlearning process
also performs the same decay at the $S/2$.
We find that the replay of learning rate scheduling
makes unlearning more stable.


\section{Experiment}






We evaluate our approach on a variety of settings including Remove A Full Category, Remove Poisoned Samples and Remove Wrongly Labeled Samples.
We also conduct in-depth analysis of the proposed method.

\paragraph{Benchmarks and Networks} 
We experiment on 4 datasets and 4 networks which results in 16 models.
Datasets we choose include CIFAR10/100 \cite{krizhevsky2009learning}, 
MNIST and Tiny-ImageNet \cite{le2015tiny}.
And networks include ResNet \cite{DBLP:conf/cvpr/HeZRS16}, 
VGG \cite{DBLP:journals/corr/SimonyanZ14a}, GoogLeNet \cite{DBLP:conf/cvpr/SzegedyLJSRAEVR15}
and DenseNet \cite{DBLP:conf/cvpr/HuangLMW17}. 

\paragraph{Hyperparameters}
As previous studies \cite{DBLP:conf/nips/MaYSCCCLQLWW21,DBLP:conf/iclr/LeH21} point out, 
different learning settings, especially the small learning rate and insufficient training epochs,
could lead to different results in network pruning. 
On CIFAR10/100 and Tiny-ImageNet, we train 160 epochs and 
the learning rate decrease by a factor of 0.1 after 80 and 120 epochs 
with initial learning rate 0.1, following \cite{DBLP:conf/nips/MaYSCCCLQLWW21}.
On MNIST, we train model at fixed learning rate 0.01 for 30 epochs.
Detailed dataset statistics and experiment setups are in Appendix \ref{sec:experiment_set}.
We set the parameter mask ratio R=$0.04$ for dataset MNIST , 
and R=$0.02$ for dataset CIFAR10/100 and Tiny-ImageNet for all mask methods (Appendix \ref{sec:remove_ratios} lists the results of different remove ratios). \footnote{In all experiment, we do not mask parameter in the final classifier.}

\paragraph{Baselines}
We compare different unlearning masking strategies (\texttt{FisherMask} in Section \ref{FisherMask} and \texttt{ActivationMask} in Section \ref{ActivationMask}
with following baselines :\footnote{
All experiments are conducted on a single 2.5GHz core and
a single NVIDIA GTX 3090 GPU.}
\begin{itemize}[leftmargin=*]
    \item \texttt{Finetune}, directly fine-tuning model $w^*$ on the remain data $D_r$ with same optimizer of the learning process.
    \item \texttt{RandomMask}, randomly masking parameters with same ratio and then fine-tuning on $D_r$.
    \item \texttt{FisherNoise}, unlearning method proposed in \cite{DBLP:conf/cvpr/GolatkarAS20} 
    which adds fisher noise to destroy the weights that may have been informative about $D_r$ (Equation \ref{eq:fishernoise2}). \footnote{
        Hyper-parameters are set as in \cite{DBLP:conf/cvpr/GolatkarAS20}.}
    For a fair comparison, we also add a fine-tuning process for \texttt{FisherNoise}.  
    \item \texttt{TF-IDF}, unlearning method proposed in \cite{DBLP:conf/www/Wang0XQ22}. 
    which uses TF-IDF score to select parameters and then fine-tunes on the dataset $D_r$.
\end{itemize}

\begin{table*}[t]
\centering
\newcommand{\std}[1]{\color{black!60}{\scriptsize $\pm$#1}}
         \resizebox{\linewidth}{!}{
         \begin{tabular}{lccc|cccc}
         \toprule
         \multirow{2}{*}{Criterion} & \multicolumn{3}{c|}{Masking $w^*$ without  Fine-tuning}  & \multicolumn{4}{c}{Masking $w^*$ with Fine-tuning}  \\ 
          \multicolumn{1}{l}{} & \emph{remain acc} (\%) & \emph{forget acc} (\%) & \emph{unlearn score}(\%) & \emph{remain acc} (\%) & \emph{forget acc} (\%) & \emph{unlearn score}(\%) & $\#$ Epochs\\ 
         \midrule
          \multicolumn{8}{l}{~ ~ \emph{ResNet20 on CIFAR10} ~ ~ ~ ~ // ~ ~ \emph{3 ($\#$ random seeds)}}\\
         \texttt{Retrain} &-&-&-&85.5\std{0.4} & 0.0\std{0.0} &  85.5\std{0.4}&133.7\std{16.1} \\
         $w^*$ (\texttt{Finetune}) & 85.0\std{0.5} & 87.6\std{0.4} &  45.3\std{0.3} &85.9\std{0.3} & 63.5\std{0.6} &  52.5\std{0.1}&4.0\std{0.0} \\
         \texttt{RandomMask} & 79.7\std{4.4} & 83.6\std{5.8} &  43.4\std{1.1} &85.6\std{0.3} & 5.0\std{2.7} &  81.6\std{1.9}&4.0\std{0.0}  \\  
         \texttt{TF-IDF} & 82.6\std{3.2} & 69.1\std{21.3} &  49.5\std{4.8} &83.2\std{3.6} & 53.8\std{10.6} &  54.2\std{1.5} & 2.7\std{1.9}\\ 
         \texttt{FisherNoise} & 78.1\std{5.0} & 0.1\std{0.1} &  78.0\std{5.0} &85.5\std{0.2} & 0.0\std{0.0} &  85.5\std{0.2}& 4.0\std{0.0}\\ 
         \texttt{ActivationMask} & 81.7\std{1.3} & 10.0\std{5.3} &  74.5\std{3.6} &85.6\std{0.5} & 0.0\std{0.0} &  85.6\std{0.5} & 4.0\std{0.0}\\
         \texttt{FisherMask} & \textbf{86.2}\std{0.4} & \textbf{0.0}\std{0.1} &  \textbf{86.1}\std{0.3}&\textbf{86.1}\std{0.4} & 0.0\std{0.0} &  \textbf{86.1}\std{0.4} &\textbf{1.0}\std{0.8}\\
         \midrule 
         
          \multicolumn{8}{l}{~ ~ \emph{GoogLeNet on CIFAR100}  ~ ~ ~ ~ // ~ ~ \emph{3 ($\#$ random seeds)}} \\
         \texttt{Retrain} &-&-&-&73.7\std{0.2} & 0.0\std{0.0} &  73.7\std{0.2} &134.7\std{8.7}\\  
         $w^*$ (\texttt{Finetune}) & 74.0\std{0.1} & 90.7\std{0.5} &  38.8\std{0.1} &73.9\std{0.2} & 75.7\std{3.8} &  42.1\std{1.0}&3.7\std{0.5} \\
         \texttt{RandomMask} & 64.3\std{1.0} & 90.7\std{5.4} &  33.8\std{1.5}&72.7\std{0.5} & 34.0\std{29.5} &  56.7\std{10.8}& 3.0\std{0.0}\\   
         \texttt{TF-IDF} & 1.0\std{0.0} & 0.0\std{0.0} &  1.0\std{0.0} &72.1\std{0.9} & 23.3\std{25.3} &  60.8\std{11.5}&3.0\std{0.8} \\ 
         \texttt{FisherNoise} & 20.5\std{3.4} & 0.0\std{0.0} &  20.5\std{3.4} &72.7\std{0.4} & 0.0\std{0.0} &  72.7\std{0.4}& 4.0\std{0.0}\\
         \texttt{ActivationMask} & 64.4\std{1.0} & 0.0\std{0.0} &  64.4\std{1.0} &73.6\std{0.3} & 0.0\std{0.0} &  73.6\std{0.3}&4.0\std{0.0} \\
         \texttt{FisherMask} & \textbf{73.8}\std{0.1} & 0.0\std{0.0} &  \textbf{73.8}\std{0.1} &\textbf{74.0}\std{0.3} & 0.0\std{0.0} &  \textbf{74.0}\std{0.3}& \textbf{2.0}\std{1.4}\\   
         \midrule



         \multicolumn{8}{l}{~ ~ \emph{Averaged on 48 runs ~ ~ ~ ~ // ~ ~ 
         4 ($\#$models) $\times$ 4 ($\#$datasets) $\times$ 3 ($\#$random seeds)}}\\
         \texttt{Retrain} &-&-&-&76.1\std{0.2} & 0.0\std{0.0} &  76.1\std{0.2}&91.7\std{6.1} \\ 
         $w^*$ (\texttt{Finetune}) &74.2\std{0.5} & 84.7\std{1.2} &  40.3\std{0.4}&74.8\std{1.2} & 54.1\std{11.0} &  50.4\std{4.0}&3.3\std{0.6} \\
         \texttt{RandomMask} &\textbf{68.5}\std{3.4} & 84.1\std{7.2} &  37.0\std{2.2} &75.8\std{0.3} & 27.2\std{7.7} &  63.1\std{3.4}& 3.5\std{0.4}\\
         \texttt{TF-IDF} &42.2\std{5.4} & 38.7\std{11.3} &  25.3\std{2.4} &75.6\std{0.5} & 37.8\std{8.2} &  58.5\std{3.7}& 3.5\std{0.5}\\
         \texttt{FisherNoise} &37.9\std{5.1} & \textbf{0.2}\std{0.1} &  37.8\std{5.1} &75.9\std{0.2} & 0.0\std{0.0} &  75.9\std{0.2}& 3.4\std{0.3} \\
         \texttt{ActivationMask} &65.6\std{3.0} & 4.4\std{2.6} &  62.6\std{4.2} &75.3\std{0.4} & 0.8\std{0.7} &  74.8\std{0.6}& 3.2\std{0.3}\\
         \texttt{FisherMask} &67.4\std{1.5} & 1.4\std{0.9} &  \textbf{66.4}\std{2.1} &\textbf{76.3}\std{0.2} & 0.0\std{0.0} &  \textbf{76.3}\std{0.3}& \textbf{2.1}\std{0.9}\\
         \bottomrule
         \end{tabular}
        }
     \smallskip
      \caption{
      Performance under different unlearning methods. The gray number in the lower right corner is the variance.
      The left part is only masking $w^*$ without fine-tuning.
      The right part is to fine-tune the masked $w^*$.
      The left half of the row for $w^*(\texttt{Finetune})$  represents $w^*$ and the right half represents \texttt{Finetune} from $w^*$.
      $\#$ Epochs denotes the number of fine-tuning epochs based on the optimal \emph{unlearn score}. 
      }
     \label{table:main_results} 
\end{table*}

\subsection{Remove A Full Category} 
Following previous works \cite{DBLP:conf/ijcai/0001IIM21,DBLP:conf/www/Wang0XQ22},
we conduct experiments on removing a full category of samples. 
Without loss of generality, we remove the first category on all datasets. 
The target unlearned model should have zero accuracy (as the same as $w_r^*$)
on the unlearn class during testing.

\paragraph{Evaluation}
We compare different methods on two parts of the test set: 
test samples belong to the unlearn category and other samples 
(accuracy on these two subsets are denoted as \emph{remain acc} and \emph{forget acc}, shortly). 
We also define a new score to characterize the overall performance of unlearning based on these two indicators: $$\emph{unlearn score} = \frac{\emph{remain acc}}{ 1 + \emph{forget acc}}.$$

Intuitively, \emph{unlearn score} is a tradeoff between  \emph{remain acc} and \emph{forget acc}.
\emph{Unlearn score} increases as  \emph{remain acc}  increases and \emph{forget acc} decreases (
$ \emph{unlearn score} = 1$ if and only if $\emph{remain acc} = 1$ and $\emph{forget acc} = 0$).
At the same time, \emph{unlearn score} penalizes very low \emph{remain score} ($ \emph{unlearn score} = 0$ if and only if $ \emph{remain acc} = 0$, no matter what value  \emph{forget acc}  takes).

\paragraph{Results}
Performance of different removing mechanisms with/without fine-tuning process on test set are listed in the Table \ref{table:main_results}.
From the results, we can find that,
\begin{itemize}[leftmargin=*]
    \item First, even without fine-tuning, masking strategies can effectively reduce \emph{forget acc}.
    Among the more effective ones are \texttt{FisherNoise},
    \texttt{ActivationMask} and \texttt{FisherMask}. 
    Although \texttt{FisherNoise} can almost perfectly remove information, it also removes too much useful information and has the lowest remain accuracy among all methods, resulting in a even lower \emph{Unlearn Score} than $w^*$.
    \texttt{FisherMask} finds a good balance between \emph{reamin acc} and \emph{foget acc}. It maintains 91\% of its original performance (67.4 vs 74.2) while almost forgot completely (1.4 vs 84.7).
    \item Second, the retrained model takes a long time to learn (needs 91 epochs to achieve the best performance averagely), 
    which indicates the necessity of unlearning methods. 
    Comparing with the retrained models, with fine-tuning process,
    all unlearning strategies could accelerate the learning process 
    and achieve a comparable performance to the final performance 
    of retrained model. Among the unlearning methods, \texttt{FisherMask} method converges faster and requires fewer fine-tuning rounds (2 epochs in average) to reach the optimal \emph{unlearn score}.
    \item Third, unlearning only with fine-tuning (\texttt{Finetune}) could be not enough. 
    The \texttt{Finetune} method could not unlearn completely 
    on most settings.
    For example, on the dataset CIFAR10/100, 
    \texttt{Finetune} method still remains high accuracy on forget set. 
    Moreover, even masking random parameters (\texttt{RandomMask}) 
    helps unlearning: it has better forget results 
    compared to \texttt{Finetune} method.
    It may because that randomly mask parameters helps optimizing
    on the new loss of the fine-tuning process, and makes it easier to 
    find a better local optimum.
    \item Finally, 
    when considering both \emph{remain acc} and \emph{forget acc},  
    \texttt{ActivationMask}, \texttt{FisherNoise} and \texttt{FisherMask} obtain appreciable increase in \emph{unlearn score}.
    Among these methods, \texttt{FisherMask} method not only unlearn completely on all experiment settings, but also exhibits the best stability among other methods, which shows the effectiveness of Fisher information in finding the key parameters. 
    \texttt{FisherNoise} performs poorly when directly masking parameters (the \emph{unlearn score} is 37.8, even lower than $w^*$), but after fine-tuning the performance is second only to \texttt{FisherMask}, which again validates the efficiency of the fisher information.
    \texttt{ActivationMask} performs comparably in average. 
    Considering a faster running time,
    \texttt{ActivationMask} can be good choice of unlearning method in most settings. 
\end{itemize}

\begin{table}[t]
\newcommand{\std}[1]{\color{black!60}{\scriptsize $\pm$#1}}
         \resizebox{\linewidth}{!}{
         \begin{tabular}{lccc}
         \toprule
         Criterion & $\Delta$(\emph{remain acc}) & $\Delta$(\emph{forget acc}) & $\Delta$(\emph{unlearn score}) \\ 
         \midrule
         \texttt{Finetune} &\textbf{1.2}\std{0.4} & 11.0\std{4.2} &  3.3\std{1.7} \\
         \texttt{RandomMask} &2.6\std{0.9} & 16.5\std{3.7} &  7.3\std{1.3} \\
         \texttt{TF-IDF} &8.9\std{1.6} & 17.7\std{3.7} &  9.2\std{1.6} \\
         \texttt{FisherNoise} &9.8\std{1.3} & \textbf{0.0}\std{0.0} &  9.9\std{1.3} \\
         \texttt{ActivationMask} &3.5\std{1.2} & 2.3\std{0.8} &  3.8\std{1.2} \\
         \texttt{FisherMask} &2.6\std{0.6} & 0.4\std{0.2} &  \textbf{2.8}\std{0.6} \\
         \bottomrule
         \end{tabular}
        }
    
     \smallskip
      \caption{Magnitude of fluctuation in test accuracy during fine-tuning process, which is calculated as $\frac{1}{S-1}\sum_{t=1}^{S}
|\text{Acc}_t-\text{Acc}_{t-1}|$, $\text{Acc}_t$ is the testing accuracy of the model at $t$-th epoch. Results are averaged on all model settings (16 models * 3 runs).
      }
     \label{table:finetune_stability} 
\end{table}

Table \ref{table:finetune_stability} explores the stability of the various methods in the fine-tune process, we calculate 
$\frac{1}{S-1}\sum_{t=1}^{S}
(|\text{Acc}_t-\text{Acc}_{t-1}|)$
to measure the stability
($\text{Acc}_t$ is the testing accuracy of the model at $t$-th epoch, $S$ is set as 5).  
Smaller scores indicate smaller fluctuations.
From the results, we can see that different methods fluctuate differently in fine-tuning process: \texttt{Finetune} and \texttt{TF-IDF}
fluctuate wildly on \emph{forget acc}. The possible reason is that without removing key information, the model is relying on catastrophic forgetting to unlearn which is less stable.
\texttt{FisherNoise} method fluctuate wildly on \emph{remain acc} which denotes it removes too much information.
\texttt{FisherMask} method fluctuates very little on all three metrics, demonstrating a good stability in the fine-tuning process. 

\subsection{Remove Poisoned Samples}

\begin{table}[t]
\centering
\newcommand{\std}[1]{\color{black!60}{\scriptsize $\pm$#1}}
    \resizebox{\linewidth}{!}{
         \begin{tabular}{lccc}
         \toprule
         Criterion & \emph{remain acc} & \emph{forget acc} & \emph{unlearn score} \\ 
         \midrule
         \texttt{Finetune} &82.9\std{3.4} & 88.2\std{16.7} &  44.3\std{2.3} \\
         
        \multicolumn{4}{l}{~ ~ \emph{Mask ratio = 0.15}} \\
        
        \texttt{RandomMask} &81.2\std{2.0} & 42.7\std{24.1} &  58.4\std{8.6} \\
        \texttt{ActivationMask} &78.5\std{4.1} & 11.2\std{3.3} &  70.8\std{5.9} \\
        \texttt{FisherMask} &\textbf{84.9}\std{0.1} & \textbf{11.0}\std{15.6} &  \textbf{77.9}\std{10.0} \\
        
        \multicolumn{4}{l}{~ ~ \emph{Mask ratio = 0.20}} \\
        \texttt{RandomMask} &77.0\std{1.6} & \textbf{0.0}\std{0.0} &  77.0\std{1.6} \\
        \texttt{ActivationMask} &\textbf{84.3}\std{0.3} & 8.0\std{4.8} &  78.2\std{3.6} \\
        \texttt{FisherMask} &82.3\std{2.8} & 2.2\std{3.1} &  \textbf{80.7}\std{5.0} \\

        \multicolumn{4}{l}{~ ~ \emph{Mask ratio = 0.25}} \\
        \texttt{RandomMask} &80.8\std{4.9} & 7.8\std{5.7} &  75.2\std{7.0} \\
        \texttt{ActivationMask} &\textbf{82.0}\std{3.3} & 39.8\std{38.2} &  62.4\std{13.7} \\
        \texttt{FisherMask} &79.9\std{1.2} & \textbf{0.5}\std{0.7} &  \textbf{79.5}\std{0.9} \\
         \bottomrule
         \end{tabular}
        }
    
     \smallskip
      \caption{Performances of different methods on unlearning poisoned samples. We add 200 poisoned samples with $4\times 4$ trigger size into training set.
     We test the model after unlearning poisoned samples on remain and forget set, which denote the test set and poisoned training samples, respectively. Results are averaged on 3 random seeds.
      }
     \label{table:poison} 
\end{table}

Different from removing a full class of samples, evaluating removing a group of random data points is challenge since $w^*$ and $w_r^*$ could be quite close.
To compare the effectiveness of unlearning methods on a random group, we conduct a backdoor erasing experiment.
Backdoor attack aims to insert a backdoor into the neural models during training process, which deceives the model into misclassifying the samples into a specific class when the backdoor is triggered. 
We poison 200 samples by setting a $4 \times 4$ image patch (as a trigger) in the lower right corner to 0 with a target label 0.

\paragraph{Evaluation}
We compare the unlearning performance on remain and forget sets, which denote the test set and poisoned training samples, respectively.
A complete unlearned model should have zero accuracy on poisoned training samples.
The \texttt{FisherNoise} and \texttt{TF-IDF} method are not included for they are designed to remove categories.
Same as before, we fine-tune each model for 5 epochs and report the best \emph{unlearn score}.

\paragraph{Results}
Table \ref{table:poison} shows the results on CIFAR10 dataset with ResNet20 model. We can find that: 
first, direct fine-tuning could not affect the accuracy of triggered samples, the victim model still maintain a high performance (88.2) on forget set, which suggests that it is not easy for the model to unlearn the triggered poisoned samples;
second, the value of the mask ratio slightly affects the performance of the model, but for different methods, the optimal \emph{unlearn score} is achieved with the mask ratio set to 0.2. Among these methods, \texttt{FisherMask} method can unlearn the poisoned samples almost completely, and the performance of the data to be retained is not too affected, resulting in obtaining the highest \emph{unlearn score};
finally, when the mask ratio is larger than 0.20, \texttt{RandomMask} and \texttt{ActivationMask} methods get a larger \emph{forget acc} which indicates the instability in the unlearning process.
\texttt{FisherMask} gets a smaller \emph{forget acc}, but there is also a slight performance degradation for the remaining data.

\begin{table}[t]
    \centering
    \newcommand{\std}[1]{\color{black!60}{$\pm$#1}}
    \resizebox{\linewidth}{!}{
    \begin{tabular}{lcccc}
    \toprule
    \multirow{2}{*}{Criterion}  &\multicolumn{3}{c}{Test Accuracy} \\ 
                             & 10\% & 30\% & 50\% \\       
    \midrule
    \texttt{Finetune}       & 83.04\std{0.36} & 80.05\std{1.03} & 76.91\std{0.08} \\
    \texttt{RandomMask}     & 83.22\std{0.47} & 80.16\std{0.68} & 77.23\std{0.49}  \\
    \texttt{ActivationMask} & 83.06\std{0.31} & 80.16\std{0.71} & 77.18\std{0.17} \\
    \texttt{FisherMask}     & \textbf{83.25}\std{0.22} & \textbf{80.36}\std{0.71}  & \textbf{77.45}\std{0.40}  \\
    \bottomrule
    \end{tabular}
    }
    \smallskip
    \caption{Wrongly labeled samples deletion experiment on CIFAR10 dataset with ResNet20. We create a noisy dataset by randomly shuffling labels of the training points to create outliers.
    The noise ratio is 10\%, 30\% and 50\%.
    We present the accuracy on test set. The test accuracy of the model without noisy training data 
    are 84.43 \textcolor{black!60}{$\pm$ 0.20}.}
    \label{table:outlier_deletion}
\end{table}

\subsection{Remove Wrongly Labeled Samples}
In addition to the experiment of removing backdoor samples, we also conduct an experiment of removing mislabeled noisy samples as a complement.
We create a noisy dataset by randomly shuffling labels of the training points to create outliers (with different noisy ratios), and then remove them with various removing mechanisms.

\paragraph{Evaluation} We conduct the experiment on CIFAR10 with ResNet20.
In this setting, we focus on the denoising capability and
report accuracy on the test set 
which characterize performances on unseen data.

\paragraph{Results} From table \ref{table:outlier_deletion}, we can find that:
first, \texttt{Finetune} has the worst performance on the test set, suggesting that simply fine-tuning can not make the model forget remembered error samples;
second, test accuracy decreases as the noise ratio increases, and \texttt{FisherMask} method achieves the highest performance at different noise ratios, indicating that it can effectively remove noise.
\texttt{ActivationMask} method has a poor performance in removing poisoned samples (see Table \ref{table:poison}) and wrongly labelled samples. We suspect the possible reason for this is that the activation values of the category information are more discriminatory, but unlearn does not work so well when the samples to be deleted are from different classes.
\texttt{RandomMask} method performs quite well when the noise ratio is small, and is comparable to \texttt{FisherMask} performance, but when the noise ratio becomes large, the performance gap becomes larger.

\subsection{More Readout Functions}\label{sec:readout_functions}

\begin{figure}
    \centering
    \includegraphics[scale=0.17]{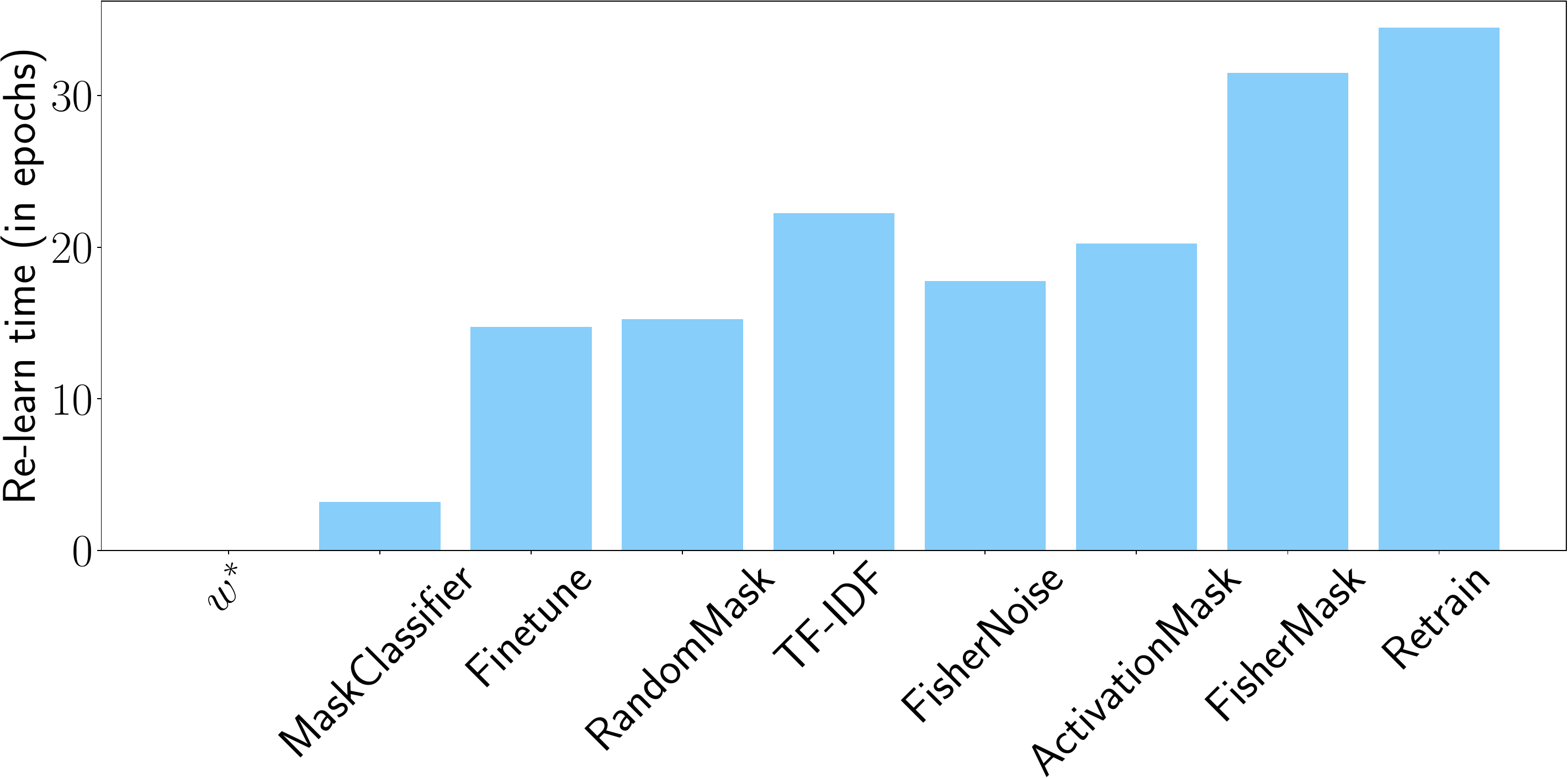}
     \caption{Re-learn time (in epochs) for removing mechanisms. 
    We count the epochs for fine-tuning the unlearned model 
    to achieve the same loss on the forget data $D_f$ 
    as $w^*$.
    \texttt{MaskClassifier} denotes masking the classifier parameters corresponding to the unlearn class.}
    \label{fig:relearn_time}
\end{figure}


Besides the prediction accuracy on different subsets,
we use a different  evaluation metric
to compare different removing mechanisms
from the perspective of internal parameters.
For convenience, the next experiments are conducted on removing a whole category.

\paragraph{Evaluation} We count the re-learn time (in epochs) for unlearned model 
to recover performance on $D_f$ (same loss) while training 
on the whole dataset $D$. 
We experiment on the CIFAR10 with ResNet20 and calculate the average epochs for each removing mechanism.
We compare with a simple baseline \texttt{MaskClassifier} 
which simply mask the classifier 
parameter corresponding to the unlearn class. 
We fix the model and only relearn the classifier parameter.
The re-learn time is ideally as the same as the model
trained without forget data $w_r^*$.
Longer re-learn time indicates the model unlearns too much information while shorter re-learn time indicates the model does not unlearn completely.

\paragraph{Results} 
Figure \ref{fig:relearn_time} shows the 
results of relearn time:
first, \texttt{MaskClassifier} can quickly 
recover the original performance which is in line with our intuition, since the model parameters still contain information about the remain set;
second, the re-learn time of \texttt{FisherMask} method is closet to $w_r^*$, and other methods have a shorter re-learn time.
The results indicate that although we can unlearn the information to make 
the accuracy of unlearn category approach zero, there still 
remains information inside the model to recover performance quickly
when original dataset is provided.


\subsection{With Limited Remain Data}

Here, we first consider the situation that the whole dataset could not be fetched and only the forget training set is available. In the previous experiments, 
we use the whole dataset for \texttt{ActivationMask} and \texttt{FisherMask} 
to find parameters to be masked. 
Here, we only have forget data for scoring parameters,
and don't run fine-tuning.
\texttt{FisherNoise} method is not included because it cannot be calculated without remain data.

Next, we consider an easier scenario where we can get a small portion of the remain data instead of the whole set. 
We randomly sample 50 samples (full remain dataset contains 45000 samples, 0.1\%) from remain training data.
We use these data both in parameter masking and fine-tuning process.

\begin{table}[t]
\centering
\newcommand{\std}[1]{\color{black!60}{\scriptsize $\pm$#1}}
    \resizebox{\linewidth}{!}{
         \begin{tabular}{lccc}
         \toprule
         Criterion & \emph{remain acc} & \emph{forget acc} & \emph{unlearn score} \\ 
         \midrule
         $w^*$ &85.0\std{0.5} & 87.6\std{0.4} &  45.3\std{0.3} \\
         
        \multicolumn{4}{l}{~ ~ \emph{No Remain Data}} \\
        \texttt{RandomMask} &\textbf{79.7}\std{4.4} & 83.6\std{5.8} &  43.4\std{1.1} \\
        \texttt{TF-IDF} &73.2\std{16.2} & 55.5\std{38.1} &  \textbf{47.4}\std{1.5} \\
        \texttt{ActivationMask} &42.0\std{21.3} & 1.0\std{0.7} &  41.5\std{21.0} \\
        \texttt{FisherMask} &45.2\std{8.8} & \textbf{0.0}\std{0.0} &  45.2\std{8.8} \\
        
        \multicolumn{4}{l}{~ ~ \emph{With 0.1\% Remain Data}} \\
        \texttt{Finetune} &79.8\std{7.1} & 75.4\std{17.6} &  45.5\std{0.6} \\
        \texttt{RandomMask} &80.0\std{2.8} & 79.3\std{8.1} &  44.7\std{0.5} \\
        \texttt{TF-IDF} &64.0\std{3.0} & 7.2\std{4.3} & 59.7\std{1.3}\\
        \texttt{FisherNoise} &84.4\std{0.5} & 0.1\std{0.0} &  84.3\std{0.4} \\
        \texttt{ActivationMask} &75.2\std{3.3} & 8.0\std{3.8} &  69.7\std{2.7} \\
        \texttt{FisherMask} &\textbf{86.1}\std{0.4} & \textbf{0.0}\std{0.0} &  \textbf{86.1}\std{0.4} \\

         \bottomrule
         \end{tabular}
        }
     \smallskip
      \caption{Performances of various unlearning methods 
          with limited fine-tuning data. 
          When no remain data is given, we only have forget data for scoring parameters and do not run fine-tuning. \texttt{FisherNoise} method is not included because it cannot be calculated without remain data.
          When we can use 0.1\% remain data, we can use these data both in parameter masking and fine-tuning process.
      }
     \label{table:only_forget}
\end{table}

Table \ref{table:only_forget} lists the results on CIFAR10 dataset with ResNet20 model. From the results, we can see that: first, when no remain training data is provided, both \texttt{ActivationMask} and \texttt{FisherMask} method can unlearn efficiently, but \texttt{ActivationMask} has a lower remain accuracy (42.0) compared to \texttt{FisherMask} (45.2). 
\texttt{TF-IDF} method has the highest \emph{unlearn score} (47.4) while \emph{forget acc} is still very high (55.5); 
second, with only 0.1\% of the remain training data provided, 
\begin{enumerate*}[label=(\roman*)]
    \item  \texttt{Finetune} and \texttt{RandomMask} still remain a high forget accuracy which shows the difficulty to complete unlearning; 
    \item \texttt{FisherNoise} method relies on the remaining data,
     and the difference in performance with the \texttt{FisherMask} method is even greater when only a small amount of data is used (84.3 vs 86.1), compared to using the full remaining data (85.5 vs 86.1).
    \item \texttt{FisherMask} can keep a stable forget performance while fully recover the remain accuracy. The remain accuracy could be boosted from  53\% (45.2 / 85.0) to  almost identical (86.1 / 85.0) with only 0.1\% of the remain data. 
    It may suggest that importance scores derived from Fisher information helps to improve data efficiency of unlearning.
\end{enumerate*}

\section{Conclusion}
In this paper, we study different masking strategies to accelerate
unlearning. 
We find our masking strategies significantly improve unlearning performance and exhibit better stability among other methods.
Experiments on various architectures and datasets show that 
all of our methods perform better than baselines and \texttt{FisherMask} method performs best while \texttt{ActivationMask} method could achieve a good performance
with a fast running speed. Future work will explore 
reducing the fine-tune time for our methods. 

%
%
%
%

\bibliography{main}

\appendix
\clearpage
\appendix
\section*{\Large \centering{Summary of the Appendix}}
This appendix contains additional details, including mathematical proofs, experimental details and additional results. The appendix is organized as follows: 
\begin{itemize}[leftmargin=*, topsep=0pt]\setlength{\parskip}{2pt}
    \item Section~\ref{sec:proof} contains proof of Proposition \ref{thm:upperbound}.
    \item Section~\ref{sec:experiment_set} lists the statistics of the datasets and training details.
    \item Section~\ref{sec:learning_rate} shows the results of experiment on different learning rate.
    \item Section~\ref{sec:remove_ratios} includes the experiment of different remove ratios of \texttt{FisherMask}.
\end{itemize}

\section{Proof of Proposition \ref{thm:upperbound}}
\label{sec:proof}

A property of Fisher matrix $F$ is that it
can be used to approximate KL-divergence between 
two distributions $p(y|x, w), p(y|x, w')$
(by Taylor expansion of 
$\mathrm{KL}(w, w')$ with respect to $w'$),
\begin{IEEEeqnarray*}{rl}
\mathrm{KL}(w, w') 
& = \mathrm{E}_{x,y}p(y|x, w)\log\frac{p(y|x, w)}{p(y|x, w')} \\
& \approx \frac{1}{2}(w-w')^TF(w-w').
\end{IEEEeqnarray*}

\begin{proof}
First we recall that 
the stationaries of $\mathcal{L}(w, D)$ satisfy normal equation,
\begin{IEEEeqnarray*}{c}
XX^Tw^* = |D|Fw^* = b, 
\end{IEEEeqnarray*}
where $X = [x_1, x_2, \ldots, x_{|D|}]$, 
$b = \sum_{i=1}^{|D|}y_ix_i$.
Denote 
$b_r = \sum_{i=1}^{|D_r|}y_ix_i$
and 
$b_f = \sum_{i=1}^{|D_f|}y_ix_i$.

We approximate KL-divergence with Fisher matrix,
\begin{IEEEeqnarray*}{rl}
\mathrm{KL}(w_r^*, \hat{w}_r)
& \approx \frac{1}{2}(w_r^*-\hat{w}_r)^TF(w_r^*-\hat{w}_r) \\
& \leq \frac{\lambda}{2|D|} \sum_j (w_{r,j}^* - \hat{w}_{r,j})^2 \\
& = \frac{\lambda}{2|D|} \left(\sum_{j\in M} {w_{r,j}^*}^2 +
\sum_{j\notin M} (w_{r,j}^* - w_j^*)^2 \right)
\end{IEEEeqnarray*}

By plugging in the weights (from normal equations),
we have,
\begin{IEEEeqnarray*}{rl}
\sum_{j\in M} {w_{r,j}^*}^2 &= 
\sum_{j\in M} \frac{1}{F_{r, jj}^2} b_{r, j}^2
\leq c_1\sum_{j\in M} \frac{1}{F_{r, jj}^2},
\\
\sum_{j\notin M} (w_{r,j}^* - w_j^*)^2 &= 
\sum_{j\notin M}
\left(\frac{1}{F_{jj}}b_i - 
\frac{1}{F_{r, jj}}b_{r, j}\right)^2 \\
& = 
\sum_{j\notin M} 
\left(\frac{1}{F_{jj}}b_{f, j} -
\frac{F_{f, jj}}{F_{jj}F_{r, jj}}b_{r, j}\right)^2 \\
& \leq 
\sum_{j\notin M} \frac{2}{F_{jj}^2}b_{f, j}^2 +
\sum_{j\notin M} \frac{2}{F_{jj}^2}
\left(\frac{F_{f, jj}}{F_{r, jj}}b_{r, j}\right)^2 \\
& \leq 
c_2\sum_{j\notin M} \frac{2}{F_{r,jj}^2} +
c_1\sum_{j\notin M} \frac{2}{F_{jj}^2}
\left(\frac{F_{f, jj}}{F_{r, jj}}\right)^2,
\end{IEEEeqnarray*}
where $c_1 = \max_j b_{r, j}^2$, $c_2 = \max_j b_{f, j}^2$.
The last inequality is based on the fact 
$F_{r, jj} \leq F_{jj}$.
Putting them together we have,
\begin{IEEEeqnarray*}{rl}
\sum_{j\in M} {w_{r,j}^*}^2 & +
\sum_{j\notin M} (w_{r,j}^* - w_j^*)^2 \\
& \leq
c_1\sum_{j\in M} \frac{1}{F_{r, jj}^2} + 
c_2\sum_{j\notin M} \frac{2}{F_{r,jj}^2} \\
& + c_1\sum_{j\notin M} \frac{2}{F_{jj}^2}
\left(\frac{F_{f, jj}}{F_{r, jj}}\right)^2 \\
& \leq c + 
2c_1\sum_{j\notin M} \frac{1}{F_{jj}^2}
\left(\frac{F_{f, jj}}{F_{r, jj}}\right)^2,
\end{IEEEeqnarray*}
where $c = \max\{c_1, 2c_2\}\sum_{j} \frac{1}{F_{r, jj}^2}$.

\end{proof}

\section{Experiment Settings}\label{sec:experiment_set}

\paragraph{Dataset and Training Statistics} 
To ensure we get the generic conclusions among datasets and structures, 
we run our unlearning strategies comparing with baseline methods 
on 4 common datasets and 4 common architectures.
We list the data statistics in Table \ref{tab:data_statistics} 
and Table \ref{tab:experiment_setup}. 
\begin{table}[h]
    \resizebox{0.95\linewidth}{!}{
    \centering
    \begin{tabular}{lcccc}
    \toprule
    Dataset  & CIFAR10 & CIFAR100 & MNIST   & Tiny-ImageNet \\
    \midrule
    \# images & 50K/10K & 50K/10K  & 60K/10K & 100K/10K      \\
    \# class  & 10      & 100      & 10      & 200           \\
    Img Size & 32*32   & 32*32    & 28*28   & 64*64        \\
    \bottomrule
    \end{tabular}
    }
    \smallskip
    \caption{Statistics on datasets.}
    \label{tab:data_statistics}
\end{table}

\begin{table}[h]
    \centering
    \resizebox{0.95\linewidth}{!}{
    \begin{tabular}{lccc}
    \toprule
    Experiments                 & CIFAR10/100   & MNIST & Tiny-ImageNet \\
    \midrule
    Training epochs             & 160           & 30    & 160           \\
    Batch size                  & 128           & 128   & 32            \\
    Init learning rate          & 0.1           & 0.1   & 0.1           \\
    Optimizer                   & SGD           & SGD   & SGD           \\
    Learning rate scheduler     & step          & N/A   & step          \\
    Learning rate decay (epoch) & {[}80, 120{]} & N/A   & {[}80, 120{]} \\
    Learning rate decay factor  & 10            & N/A   & 10            \\
    Momentum                    & 0.9           & 0.9   & 0.9           \\
    Warmup epochs               & 0             & 0     & 20            \\
    \bottomrule
\end{tabular}
    }
    \smallskip
\caption{Detailed experiment setups.}
\label{tab:experiment_setup}
\end{table}
\section{Learning Rate Scheduling}\label{sec:learning_rate}
\begin{figure*}[t]
    \centering     
    \begin{minipage}{.32\linewidth}
        \centering    
        \includegraphics[scale=0.22]{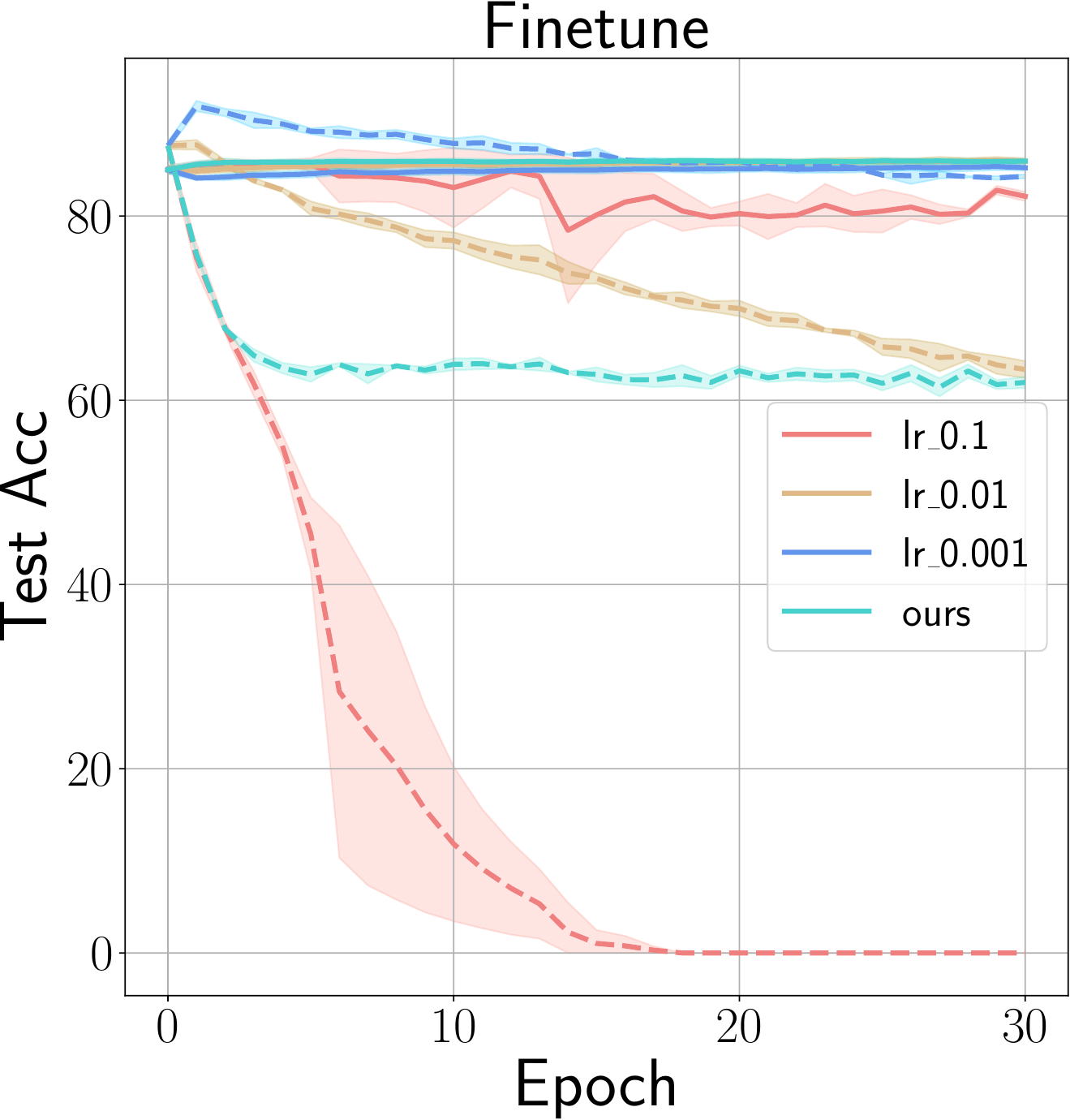} 
    \end{minipage}
    \begin{minipage}{.32\linewidth}
        \centering    
        \includegraphics[scale=0.22]{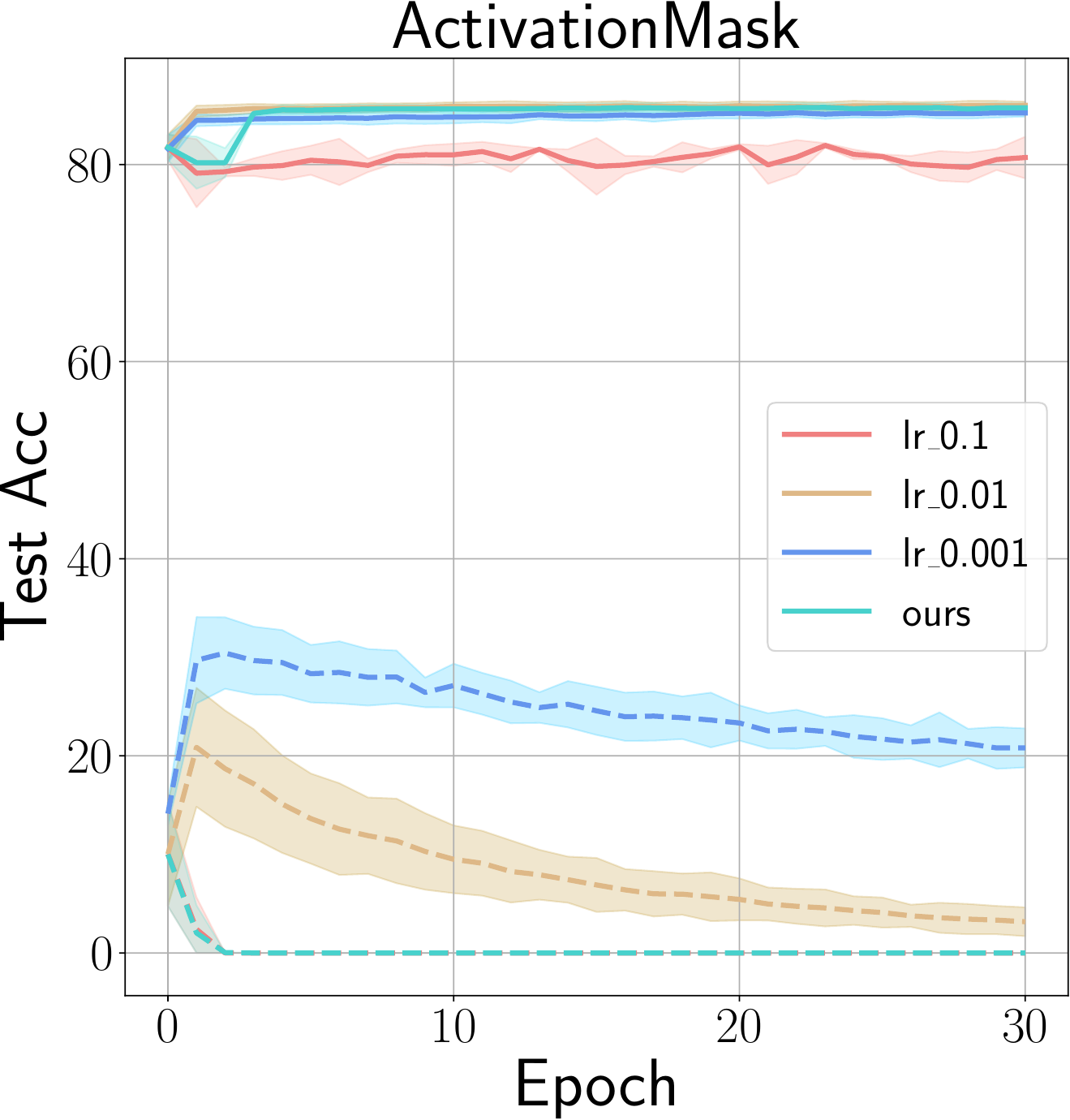}
    \end{minipage}
    \begin{minipage}{.32\linewidth}
        \centering    
        \includegraphics[scale=0.22]{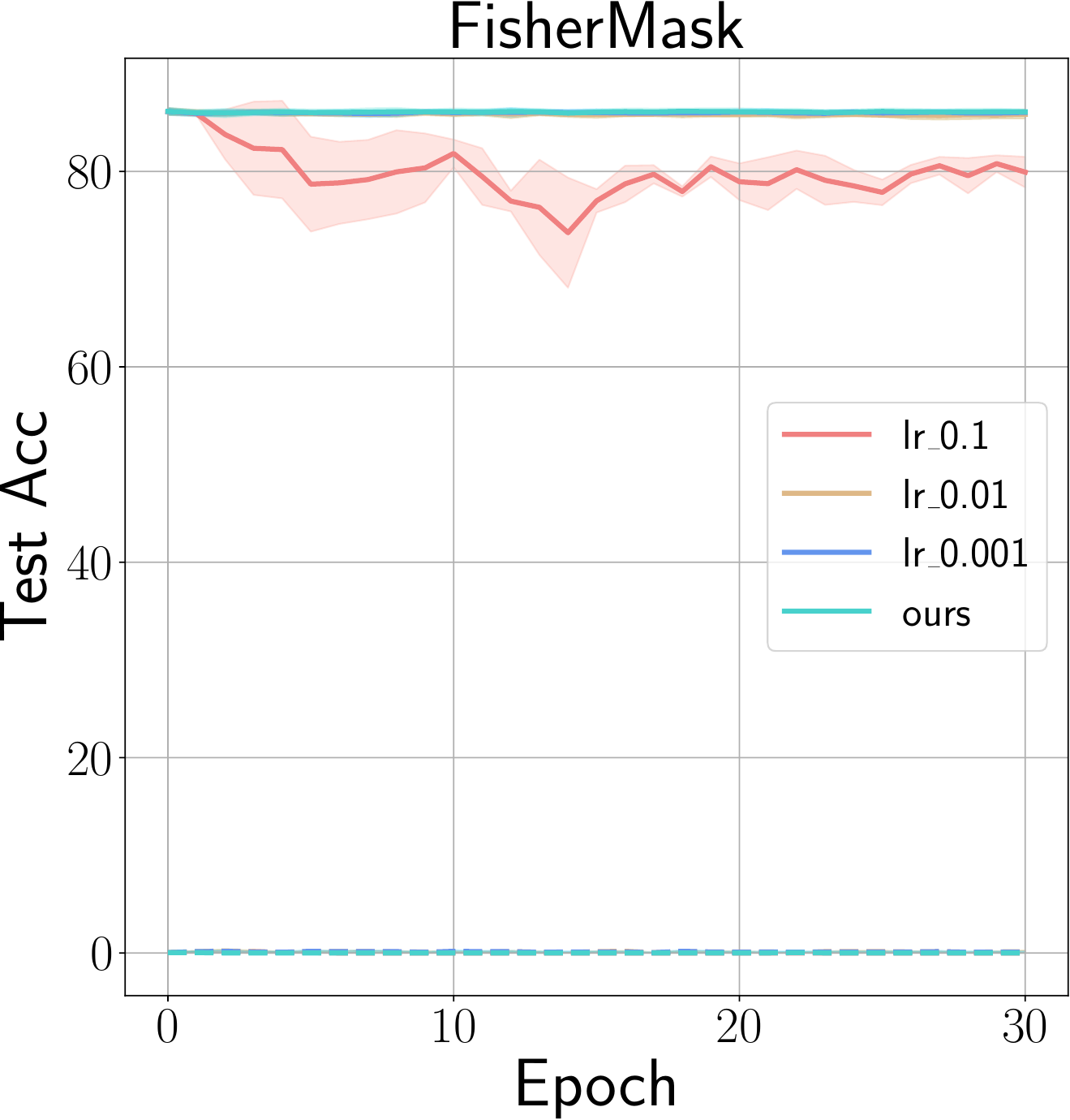}
    \end{minipage}
    \caption{Experiment results of different learning rate  schedule used in fine-tuning process on CIFAR10 dataset with ResNet20. The solid and dashed lines indicate the performance on the remain and forget datasets, respectively. }    
    \label{fig:learning_rate_exper}  
\end{figure*}

Considering that usually the learning rate starts from a large value and decays slowly during epochs, how to choose a appropriate learning rate in our fine- tuning process could be a problem. 
A large learning rate could be helped for accelerating learning process, while a small learning rate helps to approach local minima and get better performance. 
As we want to recover the performance of remain data as quickly as possible (as few fine-tuning epochs as possible), we compress the scheduler of the original learning rate to the first $S$ epochs (Section \ref{finetune}).
We show the results of different learning rate on CIFAR10 with ResNet20 model in Figure \ref{fig:learning_rate_exper}. 

It can be seen that compared to using a constant learning rate: \footnote{0.1, 0.01 and 0.001 are the learning rates at the initial time, after the first decay and after the second decay, respectively.}
\begin{enumerate*}[label=(\roman*)]
    \item for \texttt{Finetune}, a large learning rate is helpful for forgetting, but may lead to a worse and less stable remain performance. 
    A smaller learning rate helps keeping high remain accuracy but is harmful to the forgetting performance. 
    Our scheduler could find a balance between
    stable high remain accuracy and low forget accuracy.
    \item for \texttt{ActivationMask}, the results are similar.
    Our scheduler could recover the performance of the remain data as quickly as possible (as small learning rate) while maintaining stability and accelerating unlearning (as large learning rate).
    \item for \texttt{FisherMask}, it is less influenced by the learning rate.
\end{enumerate*}

\begin{figure}
    \centering
    \includegraphics[scale=0.22]{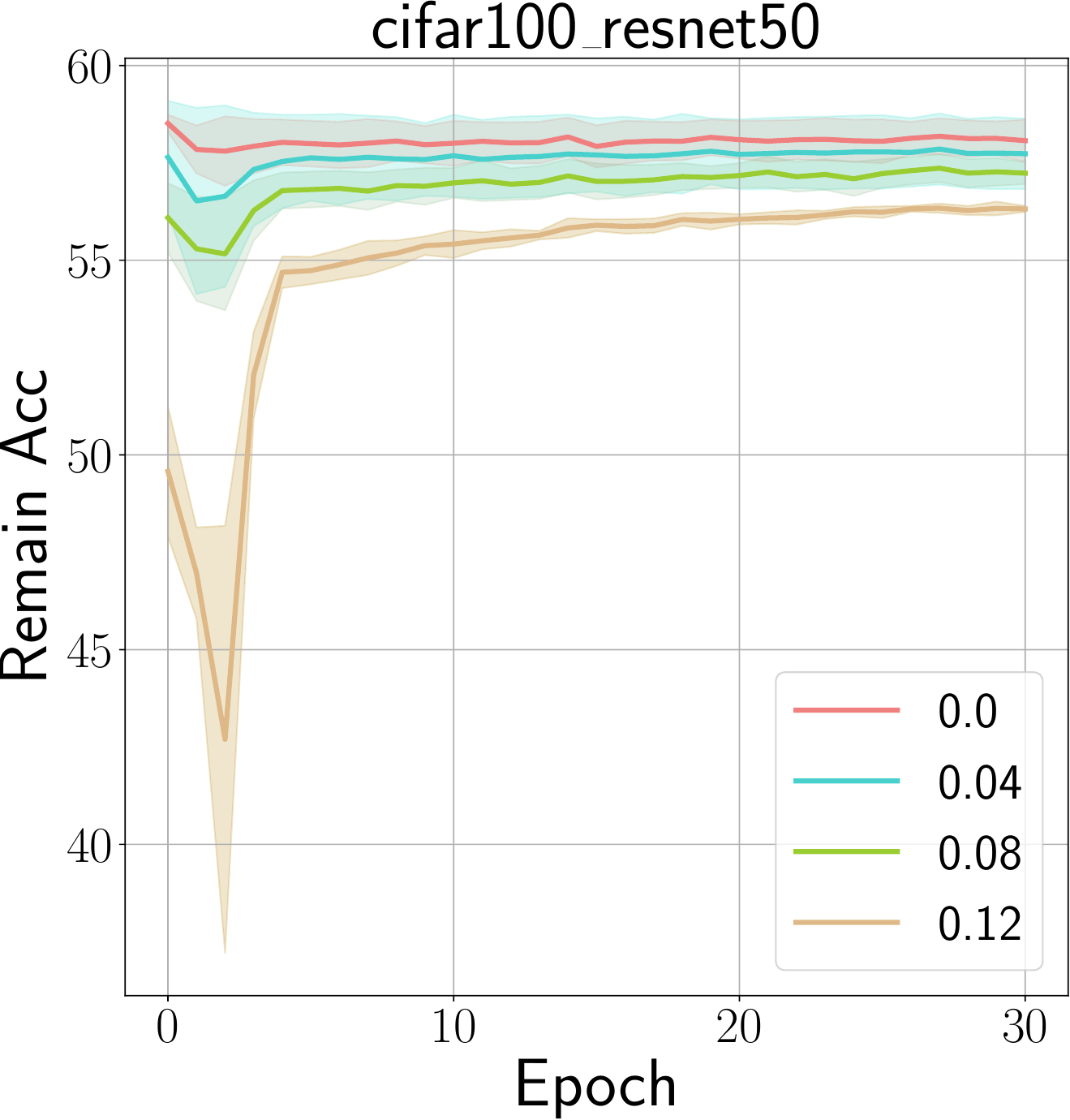}
    \caption{Performances on remain data for \texttt{FisherMask} method 
    with different removing ratios. 
    \texttt{FisherMask} could unlearn completely under all removing ratios,
    but as the remove ratios become larger, 
    the performance on remain data degrades.}
    \label{fig:remove_ratio}
\end{figure}
\section{Different Removal ratios}
\label{sec:remove_ratios}
Here we show the performance for \texttt{FisherMask} method with different remove ratios on remain dataset in Figure \ref{fig:remove_ratio}.
Accuracy on forget dataset remains 0 as remove ratio ranges from 0 to 0.12,
but accuracy on the remain set changes a lot. 
Performances degrade significantly 
as the percentage of masking increases. 
Besides that, oscillation of curves 
also becomes progressively larger as the increase of remove ratio.
When the remove ratio is relatively small, the performance change curve is relatively flat.
It will have a small drop of performances (because the learning rate is a bit large at the start)
and quickly pick up. However, the performance drops a lot when the remove ratio is higher.
Therefore, when we scrub too many information,
it is hard for fine tuning to find them back 
even with the full remain training set.
\end{document}